%% file: AIM_Slutski_Chablat_Angeles.tex
\documentclass[letterpaper,10pt,twocolumn]{article}
\usepackage{psfig}
\usepackage{epsfig}
\input{macro.tex}

\makeatletter
\def\@normalsize{\@setsize\normalsize{12pt}\xpt\@xpt
\abovedisplayskip 10pt plus2pt minus5pt\belowdisplayskip
\abovedisplayskip
\abovedisplayshortskip \z@ plus3pt\belowdisplayshortskip 6pt plus3pt minus3pt
\let\@listi\@listI}
\def\subsize{\@setsize\subsize{12pt}\xipt\@xipt}
\def\section{\@startsection {section}{1}{\z@}{24pt plus 2pt minus 2pt}
{12pt plus 2pt minus 2pt}{\large\bf}}
\def\subsection{\@startsection {subsection}{2}{\z@}{12pt plus 2pt minus 2pt}
{12pt plus 2pt minus 2pt}{\subsize\bf}}
\makeatother

\begin{document}
\date{}
\title {\Large\bf The Kinematics of Manipulators Built From Closed Planar
Mechanisms} 



 \author{ 
  Leonid Slutski$^{\ast}$, Damien Chablat$^{\circ}$ and Jorge Angeles$^{\ast}$ \\
\begin{tabular}[t]{c@{\extracolsep{5em}}c}
 \\
  $^{\ast}$Department of Mechanical Engineering & 
  $^{\circ}$INRIA Rocquencourt  \\ 
   \& Centre for Intelligent Machines & 
   Domaine de Voluceau, B.P. 105   \\
   McGill University  & 78153 Le Chesnay \\
   Montreal, QC  H3A 2A7 Canada  & 
   France \\
   {\em slutski@cim.mcgill.ca \hfill angeles@cim.mcgill.ca} &
   {\em chablat@cim.mcgill.ca}
\end{tabular}}
\maketitle



\thispagestyle{empty}
\subsection*{\centering Abstract}
{\em The paper discusses the kinematics of manipulators builts of planar closed kinematic chains. A special kinematic scheme is extracted from the array of these mechanisms that looks the most promising for the creation of different types of robotic manipulators. The structural features of this manipulator determine a number of its original properties that essentially simplify its control. These features allow the main control problems to be effectively overcome by application of the simple kinematic problems. The workspace and singular configurations of a basic planar manipulator are studied. By using a graphic simulation method, motions of the designed mechanism are examined. A prototype of this mechanism was implemented to verify the proposed approach.}
\begin{keyword}
Kinematics, Manipulator, Closed Planar Mechanism, Singularity, Workspace
\end{keyword}
\section{Introduction}
Closed kinematic chains  are promising building blocks to build 
novel and effective parallel manipulators. There are two principal directions in 
the synthesis of these machines. The first is based on the use of platform 
spatial manipulators. This approach has its 
origin in the Stewart-Gough platform \cite{stewart} and has been
studied extensively \cite{hunt}. It is well known that platform 
manipulators are characterized by high stiffness and 
accuracy, but, at the same time, have a restricted workspace and pose 
 some control difficulties because of their quite complicated 
direct kinematics.
\par
Therefore, another direction is now under development, based on the 
use of  closed planar kinematic chains as building blocks of
spatial robots. We claim that this very 
promising approach has not yet been fully exploited.
\par
The  simplest example of a basic planar manipulator of this type, shown
in Figure 1, has motivated intensive research (e.g., \cite{bajpai}).
 This mechanism is based on the use of a dyad, that is, a planar 
group of the second class, links 1 and 2, according to the classification 
of Assur-Artobolevskii \cite{art}.

Rotation about a vertical axis provides this mechanism with three-degree-of-freedom (dof)
motion capabilities. The advantages of the mechanism are enhanced stiffness and driving motor placement on the base (joints $A$ and $B$), both advantages being common properties  of parallel manipulators. However, only manipulators 
built on planar closed chains have the advantages of rather simple kinematics and a relatively large
 workspace. In fact, the layout of Fig.~1 was so effective that it has been the first closed kinematic chain used in one of the versions of the German  ``Kuka'' industrial robot.
\par
 A disadvantage of the scheme shown in Fig.~1 is its somewhat restricted workspace, determined by the distance between the base joints $A$ and $B$. Based on kinematic considerations, this distance may be chosen to be near zero, as 
was practically implemented in the design of the ``Kuka'' robot. 
\begin{figure}[h]
\centerline{
 \makebox{\psfig{figure=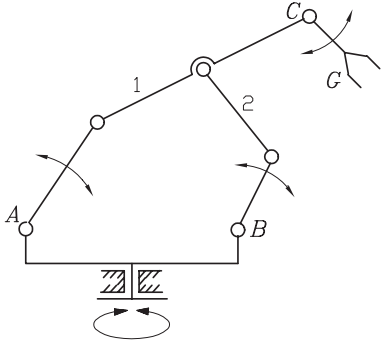,width=5.cm}} 
}
\caption{Three-dof manipulator based on planar mechanism of second
  class}
\end{figure}
Another feature of the basic mechanism (Fig.~1) is that if we need to control 
the orientation of the gripper $G$, it is usually necessary to mount an additional 
actuator in the joint $C$ of the moving link 1. 
\par
A solution that allows one to solve this problem without putting a motor on the 
moving link involves the third class group \cite{art} as a basic
kinematic chain (Fig. 2). 
\begin{figure}[h]
  \centerline{
    \begin{tabular}{c}
      \makebox{\psfig{figure=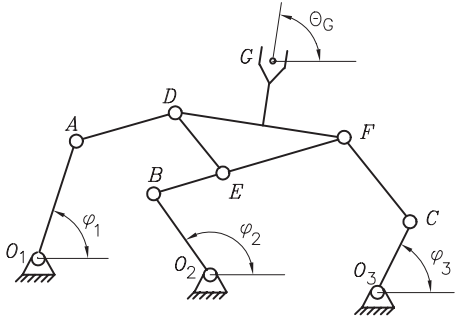,width=6.cm}} 
    \end{tabular}}
  \caption{Planar manipulator based on the mechanism of third class}
\end{figure}
This mechanism has gained its reputation thanks to
Hunt (e.g., \cite{hunt}) and the following publications
\cite{gosselin,pennok}. It is, in fact, very important and interesting because it serves as a link between the two above-mentioned approaches in  organization of robot platform mechanisms. Indeed, it is possible to pass from the  planar closed mechanism (Fig. 2) to the platform spatial manipulator by changing the revolute joints (points $D, E$ and $F$) by spherical joints and by removing the dyads $O_{1}A, AD; O_{2}B, BE;$ and $O_{3}C, CF$ onto different planes. 
\par
The third-class mechanisms are the most promising to organize prospective spatial industrial robots, as demonstrated in a  patent \cite{konst}. In this design, the moving platform, link $DEF$, of the Assur group (Fig.~2) was attached to the prismatic kinematic pair directed orthogonally to the plane of group location.
\par
Third-class mechanisms have good prospects because of a quite simple kinematics. In this connection, a number of investigations were carried out  to determine theirs workspace, singular configurations and other characteristics (\cite{gosselin,pennok}). Currently, some propositions have been made in which linear actuators of robotic mechanisms 
are used as input links \cite{herve}. 
\par
This paper develops this approach in creating closed structural schemes for platform robot mechanisms. A special variation of the discussed mechanisms with a linear platform link \cite{merlet} and other peculiarities, ensuring a high level of solution  of the manipulation tasks to be performed, is proposed for development, their design and control features being analyzed in this paper.
\section{Structure of the Three-DOF Ma\-ni\-pu\-la\-tor}
The design of a robot based on third-class chains becomes practical when the mechanism is specially constructed as discussed below. This section considers a special kind of the basic third-class planar chain, shown in the mechanism of Fig.~3.
\begin{figure}[h]
    \centerline{\hbox{
       \psfig{figure=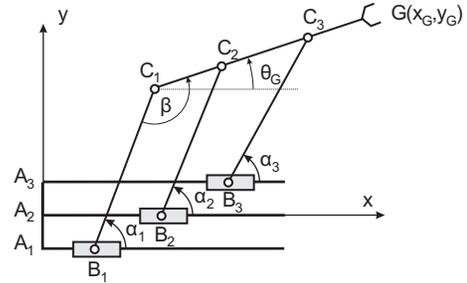,width=60mm}}}
    \caption{A scheme of the proposed base planar manipulator}
\end{figure}
Specifically, its moving platform $C_{1}G$ carries three collinear
joints,  and its actuated joints are prismatic. The actuators of these
joints  are placed in different parallel planes. The collinear form of
the  platform element prevents collisions among the links. This base manipulator performs
 spatial motions by means of additional joints with suitably-oriented axes. 
Such a structure leads to a considerable simplification of the control
as compared with the initial mechanism of Fig.~2.
\section{Kinematics}
The actuated joint variables are $\rho_1= ||A_{1}B_{1}||$, $\rho_2=
||A_{2}B_{2}||$ and $\rho_3= ||A_{3}B_{3}||$
while the Cartesian variables are the $(x_{G}, y_{G}, \theta_{G})$ coordinates
of the end-effector (Fig. 3). Lengths $L_{11}= ||B_{1}C_{1}||$, $L_{22}= ||B_{2}C_{2}||$, 
$L_{33}= ||B_{3}C_{3}||$, $L_{1G}= ||C_{1}G||$,
$L_{2G}= ||C_{1}G||$ and $L_{3G}= ||C_{1}G||$  define the geometry of this manipulator
entirely.
\par
The velocity $\dot\negr g$  of the point $G$ can be obtained in
three different forms, depending on the direction in which the loop
is traversed \cite{chablat, daniali}, namely:
\beq
        \dot{\negr g}= \dot{\negr b}_1
                     + \dot{\alpha_1} \negr E (\negr c_1 - \negr b_1)
                     + \dot{\theta_G} \negr E (\negr g - \negr c_1)
        \protect\label{equation:kinematib_1}
\eeq
\beq
        \dot{\negr g}= \dot{\negr b}_2
                     + \dot{\alpha_2} \negr E (\negr c_2 - \negr b_2)
                     + \dot{\theta_G} \negr E (\negr g - \negr c_2)
        \protect\label{equation:kinematib_2}
\eeq
\beq
        \dot{\negr g}= \dot{\negr b}_3
                     + \dot{\alpha_3} \negr E (\negr c_3 - \negr b_3)
                     + \dot{\theta_G} \negr E (\negr g - \negr c_3)
        \protect\label{equation:kinematib_3}
\eeq
with matrix ${\bf E}$ defined as
\beq
{\bf E}= \left[\begin{array}{cc}
              0 & -1 \\
              1 &  0
             \end{array}
        \right]                  \nonumber
\eeq
and $\negr b_i$ and $\negr c_i$ denoting the position vectors in the frame $x$-$y$ of Fig.~3 of the points $B_i$ and $C_i$ respectively, for $i=1, 2, 3$.
\par
Furthermore, note that vectors $\dot{\negr b}_i$ are given by
\beq
  \dot{\negr b}_i= \dot{\rho}_i \frac{{\gbf \rho_i}}{||\negr \rho_i||}.
\eeq
We would like to eliminate the three idle joint rates $\dot{\alpha}_1$, $\dot{\alpha}_2$ and $\dot{\alpha}_3$ from eqs.(\ref{equation:kinematib_1}-\ref{equation:kinematib_2}-\ref{equation:kinematib_3}), which we do upon dot-multiplying their
two sides by $\negr c_i-\negr b_i$, thus obtaining
\beq
  (\negr c_1-\negr b_1)^T \dot{\negr g}
  = (\negr c_1-\negr b_1)^T \dot{\rho}_1 \frac{{\gbf \rho_2}}{||\negr \rho_1||}
  + (\negr c_1-\negr b_1)^T \dot{\theta_G} \negr E (\negr g - \negr c_1),
  \protect\label{equation:kinematib_1bis}
\eeq
\beq
  (\negr c_2-\negr b_2)^T \dot{\negr g}
  = (\negr c_2-\negr b_2)^T \dot{\rho}_2 \frac{{\gbf \rho_2}}{||\negr \rho_2||}
  + (\negr c_2-\negr b_2)^T \dot{\theta_G} \negr E (\negr g - \negr c_2),
  \protect\label{equation:kinematib_2bis}
\eeq
\beq
  (\negr c_3-\negr b_3)^T \dot{\negr g}
  = (\negr c_3-\negr b_3)^T \dot{\rho}_3 \frac{{\gbf \rho_3}}{||\negr \rho_3||}
  + (\negr c_3-\negr b_3)^T \dot{\theta_G} \negr E (\negr g - \negr c_3).
  \protect\label{equation:kinematib_3bis}
\eeq
Equations 
(\ref{equation:kinematib_1bis}-\ref{equation:kinematib_2bis}-\ref{equation:kinematib_3bis})
can now be cast in vector form:
\beq
  {\bf A} \dot{\negr p}={\bf B \dot{\bf q}}\label{e:Adp=Bdth}
\eeq
with $\dot{\bf q}$ defined as the vector of actuated joint rates,
of components $\dot{\rho}_1$, $\dot{\rho}_2$ and $\dot{\rho}_3$ and
$\dot{\negr p}$ defined as the planar twist vector of components
$\dot{x_G}$, $\dot{y_G}$ and $\dot{\theta_G}$. Moreover ${\bf A}$ and
\negr B are, respectively, the direct-kinematics and the
inverse-kinematics matrices of the manipulator,  defined as
\begin{equation}
{\bf A}= \left[\begin{array}{cc}
                (\negr c_1 - \negr b_1)^T &
                (\negr c_1 - \negr b_1)^T \negr E (\negr g - \negr c_1)\\
                (\negr c_2 - \negr b_2)^T &
                (\negr c_2 - \negr b_2)^T \negr E (\negr g - \negr c_2)\\
                (\negr c_3 - \negr b_3)^T &
                (\negr c_3 - \negr b_3)^T \negr E (\negr g - \negr c_3)
              \end{array}
         \right]
       \protect\label{equation:jacobian_matrices_A}
\end{equation}
and
\begin{equation}
{\bf B}\!\!\!\!= \!\!\!\!\left[\begin{array}{ccc}
                \!\!(\negr c_1 \! -\! \negr b_1)^T {\gbf \rho_1} / ||\negr \rho_1||\!\!\!\!\! &
                0 &
                0 \\
                0 &
                \!\!\!\!\!\!\!\!(\negr c_2 \! -\! \negr b_2)^T {\gbf \rho_2} / ||\negr \rho_2||\!\!\!\!\!\!\!\! &
                0 \\
                0 &
                0 &
                \!\!\!\!\!(\negr c_3 \! -\! \negr b_3)^T {\gbf \rho_3} / ||\negr \rho_3||\!\!
               \end{array}
         \right] 
       \protect\label{equation:jacobian_matrices_B}
\end{equation}
\subsection{Control of Simple Motions}
An original property of the manipulator under study is its ability to carry out simple motions
either without performing any preliminary calculations, or by using some simple kinematic relationships \cite{slutski}. We summarize below these results:
\vskip 0.2cm
{\noindent \em Horizontal Translation}
\vskip 0.2cm
In this case, $\dot{y}_{G}= 0$, $\dot{\theta}_{G}= 0$ and $\dot{x}_{G}$ is arbitrary. The solution leads to a simultaneous motion of all actuators with the same velocities, that is, 
$V_{1}= V_{2}= V_{3}= V_{G}$, 
while  $V_{G}$ is the prescribed gripper velocity.
\vskip 0.2cm
{\noindent \em Vertical Translation}
\vskip 0.2cm
In this case, $\dot{x}_{G}= 0$, $\dot{\theta}_{G}= 0$,  and
$\dot{y}_{G}$ is arbitrary. Thus,
\begin{equation}
\dot{\rho_{i}}= V_{G} \tan\alpha_{i},  \; i= 1, 2, 3. 
\label{eq.8}
\end{equation}
while  $V_{G}$ is the prescribed gripper velocity.
\par
In the general case, in order to obtain the vertical end-effector
velocity, it is necessary to use the simple expressions
(\ref{eq.8}) for calculations and to measure angles $\alpha_{i}$, for $i= 1, 2, 3$.
\vskip 0.2cm
{\noindent \em Gripper Rotation}
\vskip 0.2cm
Here, $\dot{\theta}_{G}$ is arbitrary and 
$\dot{x}_{G}= \dot{y}_{G}= 0$, thus obtaining
\begin{equation}
V_{i}=  L_{iG} \dot{\theta}_{G} \frac{\sin (\alpha_{i}- \theta_{G})}{\cos \alpha_{i}}, \; i= 1, 2, 3. \label{eq.12}
\end{equation}
\par
It is apparent that the values $\alpha_{i}$ and $\theta_{G}$ have to be measured. The $\alpha_{i}$ values were already 
used for other calculations, but angle $\theta_{G}$ has to be measured only for this problem. This can be done by measuring the angle of rotation of one of the $C_{i}$  joints (Fig. 3), with the ensuring calculation of the angle  $\theta_{G}$:
\begin{displaymath}
\theta_{G}= \beta+ \alpha_{1}- 180^{\circ}.
\end{displaymath}
Thereafter, a pure rotation of the gripper can be implemented, which cannot be realized for any other design of spatial platform manipulators.
\subsection{Singular Configurations of the Proposed Manipulator}
A singularity occurs whenever \negr A or \negr B in (9) vanishes. Three types of singularities exist
\cite{gosselin}:
\beqa
    {\rm det}(\negr A) &=& 0  {\rm ~or}                    \nonumber \\
    {\rm det}(\negr B) &=& 0   {\rm ~or}                   \nonumber \\
    {\rm det}(\negr A) &=& 0 \quad {\rm and} \quad {\rm det}(\negr B) = 0.  \nonumber
\eeqa
Parallel singularities occur when the determinant of the direct
kinematics matrix \negr A vanishes. The corresponding singular
configurations are located inside the workspace. They are
particularly undesirable because the manipulator cannot resist any
force and control is lost.
\par
For the manipulator study, there are two types of parallel
singularities.
\par
The first type is reached whenever the lines $B_iC_i$ intersect (Fig. 4). In
such configurations, the manipulator cannot resist a wrench applies
at the intersecting point.
\begin{figure}[h]
        \centerline{\hbox{\psfig{figure=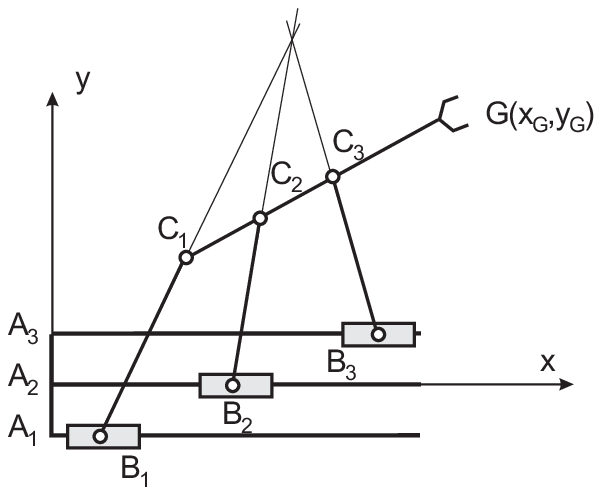,width=60mm}}}
  \caption{A parallel singularity}
\end{figure}
\par
The second type is reached whenever the lines $B_iC_i$ are
parallel (Fig.~5). That is when $(\negr c_1 - \negr b_1) \times (\negr c_2 -
\negr b_2) = \negr 0$ and $(\negr c_1 - \negr b_1) \times (\negr
c_3 - \negr b_3) = \negr 0$.
\begin{figure}[h]
        \centerline{\hbox{
   \psfig{figure=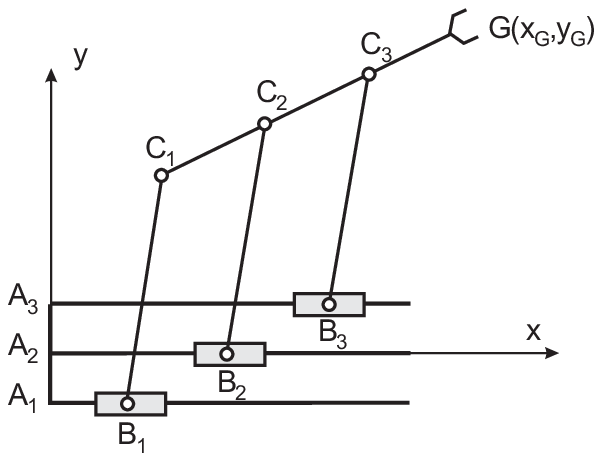,width=60mm}}}
  \caption{A parallel singularity}
\end{figure}
\par
Serial singularities occur when the determinant of the inverse kinematics matrix \negr B vanishes. When the manipulator is in such a singularity, there is a direction along which no Cartesian velocity can be produced.
\par
For the manipulator at hand,  serial singularities occur whenever at least one of the lines $A_iB_i$ is perpendicular to  $B_iC_i$, i.e 
$({\bf c_i}  - {\bf b_i})^T \gbf{\rho_i} / ||\negr
\rho_i|| = 0$, for
$i=1, 2, 3$ (Fig.~6). These singularities yield the boundary of the Cartesian workspace.
\begin{figure}[h]
        \centerline{\hbox{
   \psfig{figure=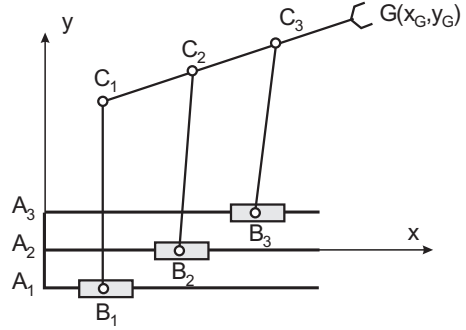,width=60mm}
}}
  \caption{A serial singularity}
\end{figure}
\section{Manipulator Workspace}
It is important to determine the manipulator workspace to exactly match its working zone. Generally,  the planar manipulator workspace is limited by a rectangle with height $h$ and width $w$. The value  $h$ may be determined as 
\begin{displaymath}
h={\rm min}\{L_{1G} + L_{11}, L_{2G} + L_{22}, L_{3G} + L_{33}\},
\end{displaymath}
where we refer to variables defined in Section~3 and Fig.~3. The value of $w$ is estimated as $w= 2h + L$, where $L$ is the length of the actuator strokes.
\par 
To study manipulator workspace properties, a special numerical procedure has been developed. According to this procedure, the space of the above-mentioned rectangle was divided  with a certain resolution into a number of points. For each of these points, a test was then done whether the mechanism with a corresponding set of parameters exists with a manipulator end-effector $G$ position at this point. If this condition is satisfied at least for one orientation of
the output link or not satisfied for all orientations of the output link, a passage to the next point of the rectangle is performed. This numerical procedure gives us the possibility to obtain not only an envelope of the manipulator working zone but also configurations of its dead points.
\par
One example of these results is displayed in Fig.~7.
\begin{figure}[h]
  \begin{center}
    \mbox{
    \psfig{figure=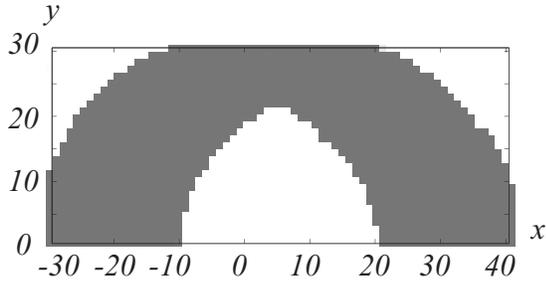,width=72mm}}
  \end{center}
\caption{Example of a manipulator workspace (a part above the $x$ axis) for
  the following set of its model
  parameters: $L_{11}= L_{22}= L_{33}= 25.0$, 
$||C_{1}C_{2}||= ||C_{2}C_{3}||= ||C_{3}G||
= 5.0, L= 10.0$
}
\end{figure}
From this graph as well as from geometric considerations, it is obvious that the value of the stroke $L$ influences the
shape of the manipulator workspace. When $L$ decreases, dead zones appear inside the manipulator envelope. This  study has been conducted and corresponding results are recorded in graph form (Fig.~8).
\begin{figure}
\centerline{
\begin{tabular}{c}
 \makebox{\psfig{figure=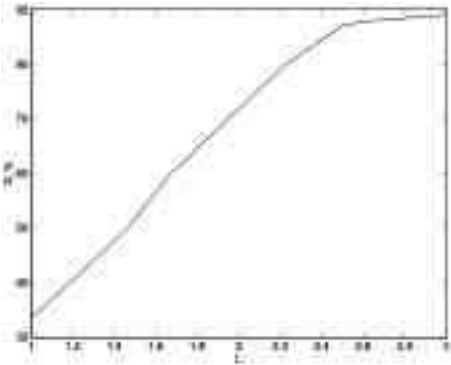,width=6.cm}} 
\end{tabular}
}
\caption{The relative value $S$ of the manipulator  workspace vs stroke length $L$}
\end{figure}
 A study was carried out for the following set of manipulator parameters: $L_{11}= 1.7, L_{22}= 1.8,
L_{33}= 1.9; ||C_{1}C_{2}||= ||C_{2}C_{3}||= ||C_{3}G|| = 1.0$; 
the value of $L$ was changed from 1 to 3. 
A characteristic of the value $S$ of the manipulator workspace was defined as a relation of the number of points, where there is at least one inverse kinematics solution, to a general quantity of the points studied  in the rectangle. 
When studying the value $S$ dependence on the stroke length $L$, one should take into account that an increase in the stroke length will lead to an increase in the workspace. However, too long stroke value may lead to a bulky mechanism. This is why, when searching for the optimal value of the stroke, it would be worth to chose it not  longer than the length allowing to exclude some dead zones inside the manipulator workspace (if there are no special requirements to manipulator performance).  Then, from the graph of Fig.~8, it may be seen that the best result is obtained for $L= 3$ (89.07\%), but a result for the value  $L= 2.5$  (87.08\%) is quite near to this maximum value.
\par
Based on these data, one may conclude that this approach
allows us to determine manipulator optimum parameters
which lead to the design of the most versatile and compact device.
\section{Simulation and Prototyping of the Proposed Manipulator}
A graphic simulation of the proposed 3D manipulator based on the
mechanism of Fig.~3 was performed by using an advanced robotics package 
on a Silicon Graphics workstation \cite{slutski}. 
Typical positioning tasks were simulated and successive spatial
motions of the robot from one location 
to another were tested. The kinematic structure was evaluated by 
animated, graphical representation of the time-varying solutions 
that includes built-in evaluation of trajectories to avoid collisions, and reachability.
One rendering of the simulation results is shown in Fig.~9.
\begin{figure}
\centerline{
  \begin{tabular}{c}
    \makebox{\epsfig{figure=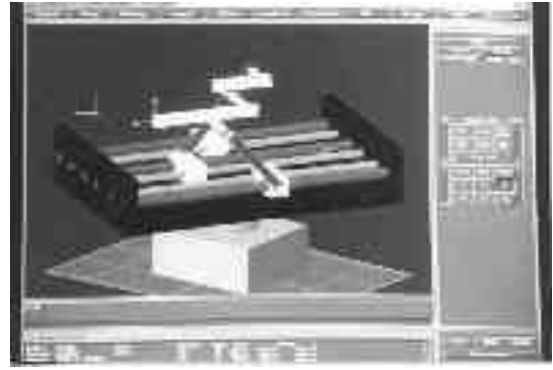,width=7.2cm}} \\
  \end{tabular}}
  \caption{Rendering of the manipulator}
\end{figure}
\par
A prototype of the planar mechanism discussed here was also built (Fig.~10) when
the first author was working at Kazakh State University (Alma-Ata,
Kazakhstan, the former USSR).
\begin{figure}
\centerline{
\begin{tabular}{c}
 \makebox{\psfig{figure=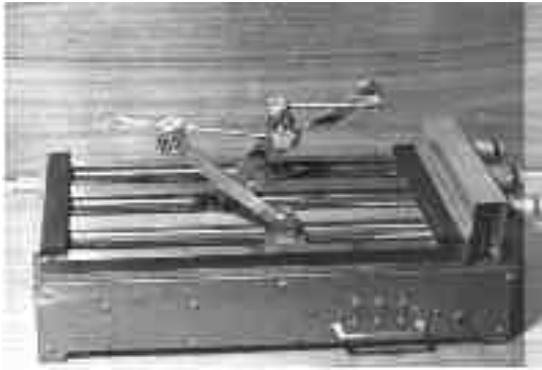,width=7.2cm}} \\
\end{tabular}
}
\caption{Manipulator prototype}
\end{figure}
The mechanism is driven by three DC motors with on-off control. The prototype allowed, for instance, to validate issues of mechanism singularities and approaches to their avoidance.
\section{Conclusions}
This paper deals with closed-chain planar mechanisms with the purpose of using them for the design of 3D parallel robotic manipulators.  The paper proposes some principles of spatial manipulator design via these mechanisms.  A paradigm is proposed that appears to be the most promising  for the design of multi-dof industrial robots. Its
peculiarities are the platform link in the form of a collinear array of attachment points and actuators that
are designed in the form of sliders placed in different parallel planes. 
\par
While using this mechanism as a basis for multi-dof manipulator design, the indicated structural features determine a number of its original properties  that essentially simplify its control. A parallel manipulator of a practical form of this design has been previously developed, which was rather similar to one  recently called the H-Robot \cite{herve}. This manipulator allows one to solve the principal control problems almost without the need to solve their inverse kinematics. In fact, the kinematic solutions are either extremely simple or do not require any calculations. For instance, the translation of the gripper along the $x$ axis (Fig.~3) may be obtained with the aid of the translation of the actuators in the required direction with equal velocities, that is, without performing any calculations. A vertical displacement of the end-effector is accomplished by moving the actuators by implementing some very simple calculations.
These special features allow one to develop rather simple control algorithms for the robot. 
\par
Workspace and singular configurations were also studied for purpose of robot design. By using graphic simulations, the motions of the designed mechanism were examined. A prototype of the discussed  mechanism was also built in order to test the proposed approach. 
\bibliographystyle{unsrt}

\end{document}

%% file: macro.tex
\newcommand{\gbf}[1] {\mbox{\boldmath${#1}$\unboldmath}}
\newcommand{\be}{\begin{equation}}
\newcommand{\ee}{\end{equation}}
\newcommand{\beq}{\begin{equation}}
\newcommand{\eeq}{\end{equation}}
\newcommand{\bed}{\begin{displaymath}}
\newcommand{\eed}{\end{displaymath}}
\newcommand{\beqa}{\begin{eqnarray}}
\newcommand{\eeqa}{\end{eqnarray}}
\newcommand{\beqann}{\begin{eqnarray*}}
\newcommand{\eeqann}{\end{eqnarray*}}
\newcommand{\bseq}{\begin{subequation}}
\newcommand{\eseq}{\end{subequation}}

\newcommand{\ba}{\begin{array}}
\newcommand{\ea}{\end{array}}

\newcommand{\negr}[1]{{\bf {#1}}}

%% file: AIM_Slutski_Chablat_Angeles.bbl
\begin{thebibliography}{99}
\bibitem{stewart} 
 D.A. Stewart, 
\newblock ``Platform with six degree of freedom,'' 
\newblock {\em Proceedings of the Institute of 
Mechanical Engineering}, 66, Vol. 180, Part 1, No. 15, pp. 371-386, 1965.

\bibitem{hunt} 
 K.H. Hunt,
\newblock``Structural kinematics of in-parallel-actuated robot-arms,''
\newblock {\em ASME Journal of Mechanisms, Transmissions, 
and Automation in Design}, Vol. 105, pp. 705-712, 1983. 

\bibitem{bajpai} 
A. Bajpai and B. Roth,
\newblock ``Workspace and mobility of a closed-loop manipulator,'' 
\newblock {\em The International 
Journal of Robotics Research}, Vol. 5, No. 2, pp. 131-142, 1986.

\bibitem{art} 
 I.I. Artobolevskii, 
\newblock {\em Theory of Mechanisms and Machines},
\newblock Nauka, Moscow, 1988 (in Russian).

\bibitem{sandor}
 G.N. Sandor and A.G. Erdman,
\newblock {\em Mechanical Design: Analysis and Synthesis},
\newblock Prentice Hall, 1984. 

\bibitem{gosselin} 
C. Gosselin and J. Angeles,
\newblock ``Singularity analysis of closed loop kinematic chains,'' 
\newblock {\em IEEE 
Transactions on Robotics and Automation}, Vol. 6, No. 3, pp. 281-290,
1990.

\bibitem{pennok} 
G. Pennok and D. Kassner,
\newblock ``The workspace of a general geometry planar three-degree-of-freedom 
platform-type manipulator,'' 
\newblock {\em ASME Journal of Mechanical Design}, Vol. 115,
pp. 269-276, 1993.
 
\bibitem{konst} 
 U.A. Djoldasbekov,  M.S. Konstantinov,  M.D. Markov, and  L.I. Slutski, 
\newblock ``Executing mechanism of a robot-manipulator,'' 
\newblock {\em USSR patent, 
author's certificate \# 1081919}, 1983 (in Russian).

\bibitem{herve} 
 J.M. Herv\'e,
\newblock ``Design of parallel manipulators via the displacement group,'' 
\newblock {\em Proceedings Ninth World Congress on the Theory of Machines and 
Mechanisms}, Vol. 3, 29 Aug. - 2 Sept., Italy, pp. 2079-2082, 1995.

\bibitem{merlet}
 J.-P. Merlet,
\newblock ``Singular configurations of parallel manipulators and Grassman
geometry,'' 
\newblock {\em The International Journal of 
Robotics Research}, Vol. 8, No. 5, pp. 45-56, 1989. 

\bibitem{chablat}
D. Chablat and Ph. Wenger, 
\newblock ``Working modes and aspects in fully-parallel manipulator'',
\newblock {\em Proceeding IEEE International Conference on Robotics
  and Automation}, 
pp. 1964-1969, May 1998. 

\bibitem{daniali} 
 H.R. Mohammadi Daniali, P. Zsombor-Murray, and J. Angeles,
\newblock ``Singular analysis of planar parallel manipulators,'' 
\newblock {\em Mechanism and 
Machine Theory}, vol. 30, no. 5, pp. 665-678, 1995. 

\bibitem{slutski}
L. Slutski, 
\newblock ``Closed plane mechanisms as a basis of parallel
 manipulators''. 
\newblock In: J. Lenar\v{c}i\v{c} and V. Parenti-Castelli (eds.),
       {\it Recent Advances in Robot Kinematics}. Kluwer Academic 
            Publishers, Dordrecht, pp. 441-450, 1996.



\end{thebibliography}
